\documentclass[twocolumn]{svjour3}          
\smartqed  
\usepackage{amssymb}
\usepackage[caption=false]{subfig}
\usepackage{tabularx}
\usepackage{multirow}
\usepackage{mathtools}
\usepackage{booktabs}
\usepackage{graphicx}
\usepackage{algorithmic}
\usepackage{algorithm}
\usepackage{url}
\usepackage{float}

\usepackage{amsthm}
\usepackage{color}
\theoremstyle{definition}

\newcommand{\etal}{\textit{et al.}}
\newcommand{\ie}{\textit{i.e.}}
\newcommand{\eg}{\textit{e.g.}}
\newcommand{\viz}{\textit{viz.}}

\newcommand{\fig}[1]{Fig. \ref{#1}}
\newcommand{\eq}[1]{eqn. (\ref{#1})}

\newcommand{\sect}[1]{Sect. \ref{#1}}
\newcommand{\tab}[1]{Table \ref{#1}}
\newcommand{\alg}[1]{Algorithm~\ref{#1}}
\newcommand{\lin}[1]{line~\ref{#1}}

\journalname{Neural Computing and Applications}

\begin{document}

\title{Hierarchical Stochastic Graphlet Embedding for Graph-based Pattern Recognition\thanks{Anjan Dutta and Pau Riba have contributed equally to this work.}}

\titlerunning{HSGE}        

\author{Anjan Dutta \and Pau Riba \and Josep Llad\'{o}s \and Alicia Forn\'{e}s}

\authorrunning{Dutta, Riba, Llad\'{o}s, Forn\'{e}s}

\institute{Anjan Dutta \at Department of Computer Science, University of Exeter, Innovation Centre, Streatham Campus, Exeter, EX4 4RN, UK \\
Tel.: +44 1392 724067 \\
\email{A.Dutta@exeter.ac.uk}
\and Pau Riba \and Josep Llad\'{o}s \and Alicia Forn\'{e}s \at Computer Vision Center, Autonomous University of Barcelona, Edifici O, Campus UAB, Bellaterra, 08193 Barcelona, Spain \\
Tel.: +34 93 581 18 28 \\
Fax: +34 93 581 16 70 \\
\email{\{priba,josep,afornes\}@cvc.uab.es}}

\date{Received: 1 August 2019 / Accepted: 22 November 2019}

\maketitle

\begin{abstract}

Despite being very successful within the pattern recognition and machine learning community, graph-based methods are often unusable because of the lack of mathematical operations defined in graph domain. Graph embedding, which maps graphs to a vectorial space, has been proposed as a way to tackle these difficulties enabling the use of standard machine learning techniques. However, it is well known that graph embedding functions usually suffer from the loss of structural information. In this paper, we consider the hierarchical structure of a graph as a way to mitigate this loss of information. The hierarchical structure is constructed by topologically clustering the graph nodes, and considering each cluster as a node in the upper hierarchical level. Once this hierarchical structure is constructed, we consider several configurations to define the mapping into a vector space given a classical graph embedding, in particular, we propose to make use of the Stochastic Graphlet Embedding (SGE). Broadly speaking, SGE produces a distribution of uniformly sampled low to high order graphlets as a way to embed graphs into the vector space. In what follows, the coarse-to-fine structure of a graph hierarchy and the statistics fetched by the SGE complements each other and includes important structural information with varied contexts. Altogether, these two techniques substantially cope with the usual information loss involved in graph embedding techniques, obtaining a more robust graph representation. This fact has been corroborated through a detailed experimental evaluation on various benchmark graph datasets, where we outperform the state-of-the-art methods.

\end{abstract}

\section{Introduction}
\label{sec:intro}
Graph-based methods have been very successful for pattern recognition, computer vision and machine learning tasks~\cite{Conte2004,Foggia2014,Vento2015}. However, due to their symbolic and relational nature, graphs have some limitations if we compare them with the traditional statistical (vector-based) representations. Some trivial mathematical operations do not have an equivalence in the graph domain. For example, computing pairwise sums or products (which are elementary operations in many classification and clustering algorithms) is not defined in a standard way in the graph domain. In the literature, a possible way this problem has been addressed is by means of embedding functions. Given a graph space $\mathbb{G}$, an \emph{explicit embedding} function is defined as $\varphi:\mathbb{G}\rightarrow \mathbb{R}^n$ which maps a given graph to a vector representation~\cite{Bunke2010,Gibert2012,Luqman2013,Saund2013,Shervashidze2009a} whereas an \textit{implicit embedding} function is defined as $\varphi:\mathbb{G}\rightarrow \mathcal{H}$ which maps a given graph to a high dimensional Hilbert space $\mathcal{H}$ where a dot product defines the similarity between two graphs $K(G,G')=\langle \varphi(G),\varphi(G') \rangle$, $G,G'\in\mathbb{G}$~\cite{Dupe2010,Gartner2003,Horvath2004,Kashima2004}. In the graph domain, the process of implicitly embedding graph is termed as \emph{graph kernel} which basically defines a way to compute the similarity between two graphs. However, defining such embedding functions is extremely challenging, when the constraints on time efficiency and preserving the underlying structural information is concerned. The problem becomes even more difficult with the growing size of graphs, as the structural complexity increases the possibility of noise and distortion in structure, and raises risk of loosing information. Hierarchical representation is often used as a way to deal with noise and distortion~\cite{Mousavi2017,Ulrich2012}, which provides a stable delineation for an underlying object. Hierarchical representations allow to incrementally contract the graph, in a space-scale representation, so the salient features (relevant subgraphs) remain in the hierarchy. Thus, top levels become a compact and stable summarization.

Processing information using a multiscale representation is successfully employed in computer vision and image processing algorithms, which is mostly inspired by its resemblance with human visual perception~\cite{Adelson1984}. It is observed that a naturalistic visual interpretation always demands a data structure able to represent scattered local information as well as summarized global facts~\cite{Jolion1994}. Hierarchical representation is often used as a paradigm to efficiently extract the global information from the local features. Apart from that, hierarchical models are also believed to provide time and space efficient solutions~\cite{Ulrich2012}. Motivated by the above mentioned intuition and the existing works in the related fields, many authors have come up with different hierarchical graph structures for solving various problems~\cite{Farabet2013,Fei2005,Marfil2007,Ulrich2012}. In this sense, it is worth to mention the work of Mousavi~\etal~\cite{Mousavi2017}, who presented a hierarchical framework for graph embedding, although they did not explore the complex encoding of the hierarchy.

In this paper, motivated by the successes of the hierarchical models and the efficiency of graph embedding theory, we propose a general hierarchical graph embedding formulation that first creates a hierarchical structure from a given graph, and then utilizes the multiscale structure to explicitly embed a graph in a real vector space by means of local graphlets. First, we make use of the graph clustering algorithm proposed in~\cite{Girvan11062002} to obtain a hierarchical graph representation of a given input graph. Here, each cluster of nodes in a level $i$ is depicted as a single node in the upper hierarchical level $i+1$, whereas the edges in a level are connected depending on the original topology of the base graph, and the hierarchical edges are created by joining a node representing a cluster to all the nodes in the lower level. Thus we propose a richer encoding than Mousavi~\cite{Mousavi2017}, because our hierarchy not only contains different graph abstractions but also encodes useful hierarchical contractions through the hierarchical edges.

Once the hierarchical structure of a graph is created, we propose a novel use of the \emph{Stochastic Graphlet Embedding} (SGE)~\cite{Dutta2017a} to exploit this hierarchical information. On the one hand, we can exploit the local configuration in form of graphlets thanks to the SGE design, because graphlets provide information at different neighborhood sizes. On the other hand, the hierarchical connections allow to encode more abstract information and hence to deal with noise present in the data. As a result, the Hierarchical Stochastic Graphlet Embedding (HSGE) encodes a global and compact representation of the graph that is embedded in a vector space. The consideration of the entire graph hierarchy for the embedding instead of only the base graph empowers the representation ability and handles the loss of information that usually occurs in graph embedding methods. Moreover, the statistics obtained from the uniformly sampled graphlets of increasing size model the complex interactions among different object parts represented as graph nodes. Here, the hierarchical graph structure and the statistics of increasing sized graphlets fetch important structural information of varied contexts.

As a result, our approach produces robust representations that can benefit from the advantages of the two above mentioned strategies: we first take advantage of the embedding ability for mapping symbolic relational representations to n-dimensional spaces, so machine learning approaches can be used; and second, the ability of hierarchical structures to reduce noise and distortion inherently involved in graph representations of real data, keeping the more stable and relevant substructures in a compact way.

In conclusion, the main contribution of our work is the exploitation of the hierarchical structure of a given graph, rather than only studying the base graph for graph embedding purposes. Assessing the hierarchical information of a graph pyramid allows to extend the representation power of the embedded graph and tolerate the instability caused due to noise and distortion. Our proposal is robust because, on the one hand, it organizes the structural information in the hierarchical abstraction, and on the other hand, it considers the relation between object parts and their complex interactions with the help of uniformly sampled graphlets of unbounded size. Additionally, the proposed method is generic and can adapt any other graph embedding algorithm in the framework. In this sense, we extensively validated our proposed algorithm on many different benchmark graph datasets coming from different application domains.

The rest of this paper is organized as follows:~\sect{sec:related} describes the related works in the literature. In~\sect{sec:defn}, we introduce some definitions and notations related to the work. Our generic hierarchical graph representation is presented in~\sect{sec:hsge}. \sect{sec:sge} introduces the Stochastic Graphlet Embedding as the base embedding we will use. Afterward,~\sect{sec:expt} reports our experimental validation and compares the proposed method with available state-of-the-art algorithms. Finally, in~\sect{sec:concl} we draw the conclusions and describe the future direction of the present work.

\section{Related work}
\label{sec:related}
In what follows, we review the related works respectively on explicit and implicit graph embedding techniques, different hierarchical models and graph summarization methods, which we believed to be relevant to the main focus of the present paper.

\subsection{Graph embedding}
\label{sec:emb_ker}
Graph embedding methods are mainly divided into two different categories: (1) explicit graph embedding, (2) implicit graph embedding or graph kernel.
\subsubsection{Explicit graph embedding}
\label{sec:exp-emb}
Explicit graph embedding refers to those techniques that aim to explicitly map graphs to vector spaces. The methods belonging to this category can be further divided into four different classes. The first one, known as \emph{graph probing}~\cite{Luqman2013}, needs measuring the frequency of specific substructures (that capture content and topology) into graphs. Based on different graph substructures (\eg, node, edge, subgraph etc.) considered, different embedding techniques have been proposed. For example, Shervashidze~\etal~\cite{Shervashidze2009a} studied the non-isomorphic graphlets, albeit, node label and edge relation statistics are considered by Gibert~\etal~\cite{Gibert2012}. Saund in~\cite{Saund2013}, introduced a bottom up graph lattice in order to efficiently extract the subgraph features in preprocessed administrative documents, while Dutta and Sahbi~\cite{Dutta2017a} proposed a distribution of stochastic graphlets for embedding graphs into a vector space. The second class of graph embedding techniques is based on \emph{spectral graph theory}~\cite{Caelli2004,Wilson2005,RoblesKelly2007,Kondor2008,Kondor2009,Jouili2010}, which aims to analyze the structural properties of graphs in terms of the  eigenvectors/eigenvalues of the adjacency or Laplacian matrices of a graph~\cite{Wilson2005}. Recently, Verma and Zhang~\cite{Verma2017} proposed a family of graph spectral distances for robust graph feature representation. Despite of their relative successes, spectral methods are quite prone to structural noise and distortions. The third class of methods is inspired by \emph{dissimilarity measurements} proposed in~\cite{Pekalska2005}; in this context, Bunke and Riesen have presented several works on the vectorial description of a given graph by its distances to a number of pre-selected prototype graphs~\cite{Riesen2007a,Riesen2009a,Bunke2010,Borzeshi2013}. Motivated by the recent advancements of deep learning and neural networks, many researchers have proposed to utilize neural network for obtaining a vectorial representation of graphs~\cite{Atwood2016,Defferrard2016,Kipf2016,Niepert2016,Gilmer2017}, which results in the fourth category of methods, called \emph{geometric deep learning}.

\subsubsection{Implicit graph embedding}
\label{sec:imp-emb}
Implicit graph embedding or graph kernel methods is primarily another way to embed graphs into a vector space. They are also popular for the ability to efficiently extend the existing machine learning algorithms to non-linear data, such as, graphs, strings etc. Graph kernel methods can be roughly divided into three different categories. The first one, known as \emph{diffusion kernel}, is based on the similarity measures among the subparts of two graphs, and propagating them on the entire structure to obtain global similarity measure for two graphs~\cite{Smola2003,Lafferty2005}. The second class of methods, called as \emph{convolution kernel}, aims to measure the similarity of composite objects (modeled with graph) from the similarity of their parts (\ie~nodes)~\cite{Watkins1999}. This type of graph kernel derives the similarity between two graphs $G$, $G'$ from the sum, over all decompositions, of the similarity products of the subparts of $G$ and $G'$~\cite{Neuhaus2007}. Recently, Kondor and Pan~\cite{Kondor2016} proposed multiscale Laplacian graph kernel having the property of lifting a base kernel defined on the vertices of two graphs to a kernel between graphs. The third class of methods is based on the analysis of the common substructures that belong to both graphs, and is termed as \emph{substructure kernel}. This family includes the graph kernel methods that consider random walks~\cite{Gartner2003,Vishwanathan2010}, backtrackless walks~\cite{Aziz2013}, shortest paths~\cite{Borgwardt2005}, subtrees~\cite{Shervashidze2009a}, graphlets~\cite{Shervashidze2009} as the substructure. Different from the above three categories, Shervashidze~\etal~\cite{Shervashidze2011} proposed a family of efficient graph kernels on the Weisfeiler-Lehman test of graph isomorphism, which maps the original graph to a sequence of graphs. More recently, inspired by the successes of deep learning, Yanardag and Viswanathan~\cite{Yanardag2015} presented a unified framework to learn latent representations of substructures for graphs. They claimed that given a pre-computed kernel of graphs, their proposed technique produces an improved representation that leverages hidden representations of sub-structures.

\subsection{Hierarchical graph representation}
\label{sec:hier_mod}
In general, hierarchical models have been successfully employed in many different domains within the computer vision and image processing field, such as, image segmentation~\cite{Marfil2007,Farabet2013}, scene categorization~\cite{Fei2005}, action recognition~\cite{Niebles2007}, shape classification~\cite{Dupe2010}, graphic recognition~\cite{Broelemann2012}, 3D object recognition~\cite{Ulrich2012} etc. These approaches usually exploit some kind of pyramidal structure containing information at various resolutions. Usually, at the finest level of the pyramid, the captured information is related to local features, whereas, at coarser levels, global aspects of the underlying data are represented. This way of representation helps to interpret knowledge in a naturalistic way~\cite{Jolion1994}.

Inspired by the above intuition, hierarchical structures are often employed to extract coarse-to-fine information from a graph representation. Pelillo~\etal~\cite{Pelillo1999} proposed to match two hierarchical structures as a clique detection problem on their association graph, which was solved with a dynamic programming approach. In~\cite{Shokoufandeh2005}, Shokoufandeh~\etal~presented a spectral characterization based framework for indexing hierarchical structures that embed the topological information of a directed acyclic graph. Hierarchical representation of objects and an elastic matching procedure are also proposed from deformable shape matching in~\cite{Felzenszwalb2007}. In~\cite{Liu2008}, Liu~\etal~utilized hierarchical graph representation and a stochastic sampling strategy for layered shape matching and registration problem. A graph kernel based on hierarchical bag-of-paths where each path is associated to a hierarchy encoding successive simplifications is presented in~\cite{Dupe2010}.
Ahuja and Todorovic~\cite{Ahuja2010} used a hierarchical graph of segmented regions for object recognition. Motivated by them, Broelemann~\etal~\cite{Broelemann2012,Broelemann2013} proposed two closely related approaches based on hierarchical graph for error tolerant matching of graphical symbols. Mousavi~\etal~\cite{Mousavi2017} proposed a graph embedding strategy based on hierarchical graph representation, which considers different levels of a graph pyramid. They claimed that the proposed framework is generic enough to incorporate any kind of graph embedding technique. However, the authors did not take advantage of the complex and rich encoding of hierarchy.

From the literature review we can conclude that although there are some works in the graph domain exploiting the hierarchical graph structure, most of them are focused on some kind of error tolerance or elastic matching. Utilization of this type of multiscale representation of graph for vector space embedding is quite rare and has not been properly explored yet. This fact has worked as our motivation to work on a graph hierarchical structure for explicit graph embedding task.

\section{Definitions and notations}
\label{sec:defn}
In this section, we introduce some definitions and notations, which are relevant to the proposed work.

\begin{definition}[Attributed graph]
An \emph{attributed graph} is a $4-\text{tuple}$ $G=(V,E,L_V,L_E)$ comprising a set $V$ of \emph{vertices} together with a set $E\subseteq V\times V$ of \emph{edges} and two \emph{mappings} $L_V:V\rightarrow \mathbb{R}^m$ and $L_E:E\rightarrow \mathbb{R}^n$ which respectively assign attributes to the nodes and edges.
\end{definition}
\noindent Attributed graphs have been widely used for all sort of real world problems. The most common methodologies are error tolerant graph matching~\cite{Serratosa2000,Neuhaus2004}, graph kernels and embedding techniques~\cite{Kriege2012}.
\begin{definition}[Subgraph]
Given an attributed graph $G=(V,E,L_V,L_E)$, another attributed graph \(G'=(V',E',L_V',L_E')\) is said to be a \emph{subgraph} of \(G\) and is denoted by \(G'\subseteq G\) iff,
\begin{itemize}
    \item \(V'\subseteq V\)
    \item \(E'=E\cap V'\times V'\)
    \item \(L_V'(u)=L_V(u)\), \(\forall u \in V'\)
    \item \(L_E'(e)=L_E(e)\), \(\forall e \in E'\)
\end{itemize}
\end{definition}
\noindent A \emph{graphlet} \(g\) of \(G\) is nothing but a subgraph which inherits the topology and the attributes of \(G\). In the literature, subgraphs are often used for error tolerant matching~\cite{Suh2015,Schellewald2005,Solnon2010,LeBodic2012,Dutta2017} and frequent pattern discovery problems~\cite{Kuramochi2001,Barbu2005a,Ahuja2010}.

\begin{definition}[Hierarchical graph]
A \emph{hierarchical} \allowbreak \emph{graph} $H$ is defined as a 6-tuple $H=(V,E_N,E_H,L_V,$ $L_{E_N},L_{E_H})$ where $V$ is the set of nodes; $E_N \subseteq V\times V$ are the neighborhood edges; $E_H \subseteq V\times V$ are the hierarchical edges; \(\operatorname{L_V}\), \(\operatorname{L_{E_N}}\) and \(\operatorname{L_{E_H}}\) are three labeling functions defined as \(\operatorname{L_V}:V \rightarrow \Sigma_V \times A^k_V\), \(\operatorname{L_{E_N}}: E_N \rightarrow \Sigma_{E_N} \times A^l_{E_N}\) and \(\operatorname{L_{E_H}}: E_H \rightarrow \Sigma_{E_H} \times A^m_{E_H}\), where \(\Sigma_V\), \(\Sigma_{E_N}\) and \(\Sigma_{E_H}\) are three sets of symbolic labels for vertices and edges, \(A_V\), \(A_{E_N}\) and \(A_{E_H}\) are three sets of attributes for vertices and edges, respectively, and \(k,l,m\in \mathbb{N}\).
\end{definition}

\noindent Prior works used hierarchical structures for allowing a reasonable tolerance in the representation paradigm~\cite{Felzenszwalb2007,Dupe2010,Broelemann2013} and also for bringing robustness in the feature representation~\cite{Liu2008}.

\section{Hierarchical embedding}
\label{sec:hsge}

In the literature, only few embedding approaches exploit the idea of multiscale or abstraction information \cite{Kondor2016}. This section is devoted to provide a framework able to include this information given a graph embedding. Some works that have been proposed to exploit the mentioned multiscale information in the literature \cite{Mousavi2017,riba2017error,duttapyramidal}, discard the hierarchical information provided by the hierarchical edges and focus on abstractions of the original graph.

\subsection{Graph clustering}
\label{sssec:clustering}

\emph{Graph clustering} has been widely used in several fields such as social and biological networks~\cite{Girvan11062002}, recommendation systems~\cite{gentile2017context,li2019improved} etc. It can be roughly described as the task of grouping graph nodes into clusters depending on the graph structure. Ideally, the grouping should be performed in such a way that intra-cluster nodes are densely connected whereas the connections among inter-cluster nodes are sparse. For example, Girvan and Newman~\cite{Girvan11062002} propose a graph clustering algorithm to detect a community structures for studying social and biological networks.  Li~\etal~\cite{Li2016collaborative,Korda2016distributed,li2019improved,gentile2017context} have proposed several graph clustering techniques for recommendation systems based on different strategies: context awareness~\cite{gentile2017context}, inclusion of frequency property~\cite{li2019improved}, distributed clustering confidence~\cite{Korda2016distributed}, etc. Here we do not further review on graph clustering algorithms since it is not within the main scope of this paper. However, we would like to remark that one of the most important aspects of graph clustering is the evaluation of cluster quality, which is crucial not only to measure the effectiveness of clustering algorithms, but also to give insights on the dynamics of relationships in a given graph. For a detailed overview on effective graph clustering metrics, the interested readers are referred to~\cite{Helio2011metrics}.

Even though any graph clustering algorithm can be used, we use the standard divisive-based \emph{Girvan-Newman} algorithm~\cite{Girvan11062002} algorithm for our purpose, because it provides structurally meaningful clusters of a given graph. The \emph{Girvan-Newman} algorithm is an intuitive and well-known algorithm used for community detection in complex systems. It is a global divisive algorithm which removes the appropriate edge iteratively until all the edges are deleted. At each iteration, new clusters can emerge by means of connected components. The idea is that the edges with higher centrality are the candidates to be connecting two clusters. Therefore, \emph{betweenness centrality} measure of the edges~\cite{freeman1977set} is used to decide which edge is being removed. \emph{Betweenness centrality} on an edge $e \in E$ is defined as the number of shortest walks between any pair of nodes that cross $e$. The output of this algorithm is a dendrogram codifying a hierarchical clustering of nodes. This algorithm consists of 4 steps:

\begin{enumerate}
	\item Calculate the betweenness centrality for all edges in the network.
	\item Remove the edge with highest betweenness and generate a cluster for each connected component.
	\item Recalculate betweennesses for all edges affected by the removal.
	\item Repeat from step 2 until no edges remain.
\end{enumerate}

In this work, \emph{Girvan-Newman} algorithm is early stopped given a reduction ratio \(r \in \mathbb{R}\). Therefore, the number of clusters is forced to be \(\lfloor r \cdot |V| \rfloor\).

\subsection{Hierarchical construction}
\label{sssec:construction}

Given a graph \(G\) and a clustering \(C = \{C_1,\ldots,C_k\}\), each cluster is summarized into a new node with a representative label (see~\lin{alg:pyr_graph_cons:considervertex}). Let us consider that this label can be defined as the result of an embedding function applied to the subgraph defined by the clustered nodes and their edges. Moreover, edges between the new nodes are created depending on a connection ratio between clusters. That means that an edge is only created if there are enough connections between the set of nodes defined by both clusters (see~\lin{alg:pyr_graph_cons:ratio}). Finally, hierarchical edges are created connecting the new node \(v_{C_i}\) with all the nodes belonging to the summarized cluster \(C_i\) (see~\lin{alg:pyr_graph_cons:hierarchy1}). The proposed hierarchical construction is similar to the one proposed by Mousavi~\etal~\cite{Mousavi2017} but including explicitly the summarization generated by the clustering algorithm by means of the hierarchical edges. Thus, the proposed hierarchical construction obtains a representation which encodes abstract information by means of the clusters while keeping the relation with the original graph.

Let us introduce some notations that will be used in the following sections. Given a graph \(G\) and a number of levels \(L\), \(H_G\) denotes their corresponding hierarchical graph computed from \(G\) with \(L\) levels. \(H_G^l\), where \(l = \{0,\ldots,L\}\) is a graph without hierarchical edges corresponding to the \(l\) level of summarization, therefore, \(H_G^0 = G\). Moreover, \(H_G^{l_1,l_2}\) where \(l_i = \{0,\ldots,L\}\) and \(l_1\leq l_2\), corresponds to the hierarchical graph compressed between levels \(l_1\) and \(l_2\). Hence, \(H_G = H_G^{0,L}\) and \(H_G^l = H_G^{l,l}\). Finally, \(H_G^{l_1} \cup H_G^{l_2}\) corresponds to the union of two graphs without hierarchical edges.

\begin{algorithm}
\caption{\textsc{Pyramidal Graph Construction}($G$)}
\label{alg:pyr_graph_cons}
\begin{algorithmic}[1]
\REQUIRE $G=(V,E)$, $L$, $\varepsilon$, $\delta$
\ENSURE $H=(V,E_N,E_H)$
\STATE $H \leftarrow G$; $G_{\text{c}} \leftarrow G$
\FOR {$i=1$ to $\mathit{L}$}
    \STATE $K = \lfloor \varepsilon \cdot |G_{\text{c}}.V| \rfloor$
    \STATE $\lbrace C_1, \ldots, C_K \rbrace \leftarrow \textsc{ClusterGraph}(G_\text{c}, K)$
    \STATE $G_{\text{n}}.V = \lbrace \textsc{ConsiderAsVertex}(C_j) : j=\lbrace 1,\ldots,K\rbrace \rbrace$ \label{alg:pyr_graph_cons:considervertex}
    \FOR {$(u,v)\in G_{\text{n}}.V \times G_{\text{n}}.V$}
        \IF{$\textsc{ConnectionRatio}(u,v) \ge \delta$} \label{alg:pyr_graph_cons:ratio}
            \STATE $G_{\text{n}}.E \leftarrow G_{\text{n}}.E \cup (u,v)$
        \ENDIF
    \ENDFOR
    \FOR {$j=1$ to $\mathit{K}$}
        \STATE $E_H \leftarrow E_H \cup \lbrace (u,w) \in G_{\text{c}}.V\times G_{\text{n}}.V: \forall u\in C_j \text{ and } w=G_{\text{n}}.V_j \rbrace$ \label{alg:pyr_graph_cons:hierarchy1}
    \ENDFOR
    \STATE $G_{\text{c}} \leftarrow G_{\text{n}}$
    \STATE $H \leftarrow \textsc{IncludeToHierarchy}( H, G_{\text{c}}, E_H)$
\ENDFOR
\end{algorithmic}
\end{algorithm}

\begin{figure}[!htb]
\centering
\subfloat[]{
\includegraphics[width=0.42\linewidth]{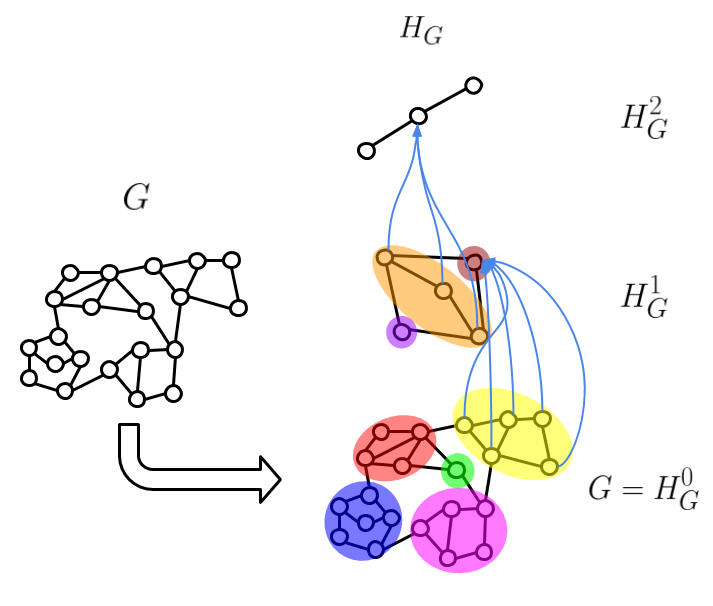}
\label{sfig:hierarchy_const}
}
\hspace{1.0cm}
\subfloat[]{
\includegraphics[width=0.38\linewidth]{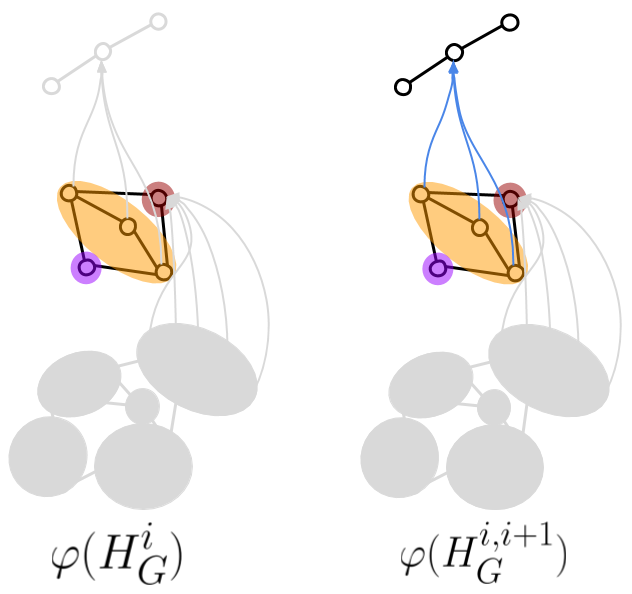}
\label{sfig:hierarchy_emb}
}
\caption{(a) Hierarchical graph construction is proposed in~\alg{alg:pyr_graph_cons}. The input graph $G$ is processed to generate a hierarchical graph $H_G$ where each level $H_G^i$ encodes a new abstraction of the original graph. Moreover, hierarchical edges provide the insights of the performed contraction. In this figure, not all the hierarchical edges have been drawn to make it easy to understand, and the node clustering is drawn in color. (b) Following the hierarchical graph construction in~\fig{sfig:hierarchy_const}, the graphs taken into consideration in order to construct the hierarchical embedding are shown. $\varphi(H_G^i)$ takes into account one abstraction level whereas $\varphi(H_G^{i,i+1})$ takes into consideration two of these levels and the hierarchical edges involved. (Best viewed in color)}
\end{figure}

\fig{sfig:hierarchy_const} shows the construction of the hierarchy given a graph $G$. Each level shows an abstraction of the input graph where the nodes have been reduced.

\subsection{Hierarchical embedding}
\label{ssec:hembed}

This section introduces a novel way to encode hierarchical information of a graph into an embedding. Moreover, the proposed technique is generic in the sense that can be used by any graph embedding function.

Given a graph \(G\) which should be mapped into a vectorial space and an embedding function \(\varphi:\mathbb{G}\rightarrow \mathbb{R}^n\), we first proceed to obtain hierarchical representation \(H_G\) following the proposed methodology in~\sect{sssec:construction}. Therefore, \(H_G\) has enriched the original graph with abstract information considering \(L\) levels. Finally, we propose to make use of the hierarchical information to construct a hierarchical embedding. The general form of the proposed embedding takes into account graphs at multiple scales and hierarchical relations. Thus, the embedding function does not only compactly encode the contextual information of nodes at different abstraction levels, but also it encodes the hierarchy contraction. The embedding function is defined as follows:

\begin{equation}
\label{eq1}
\begin{split}
\Phi(H_G) = [& \varphi(H_G^0),\ldots,\varphi(H_G^K), \\
            & \phi_1^1(H_G),\ldots,\phi_1^{k_1}(H_G), \\
            & \phi_2^1(H_G), \ldots, \phi_2^{k_2}(H_G) ]
\end{split}
\end{equation}
where,
\begin{equation} \label{eq1:levels}
\phi_1^k(H_G) = [ \varphi(H_G^{0,k}),\ldots,\varphi(H_G^{K-k,K})]
\end{equation}
\begin{equation} \label{eq1:union}
\phi_2^k(H_G) = [ \varphi(H_G^0 \cup \cdots \cup H_G^{k}),\ldots,\varphi(H_G^{K-k} \cup \cdots \cup H_G^K)]
\end{equation}
where \(K \leq L\) are the hierarchical levels taken into account and \(k_1,k_2 \leq K\) indicate the number of levels taken into account at the same time. Note that \(K=L\), \(k_1=K\) and \(k_2=K\) will take into account the whole hierarchy and possible combinations. From this general representation of the proposed embedding, we have evaluated some particular cases (the reader is referred to~\sect{sec:expt} for more details on the experimental evaluation).

\textbf{Baseline embedding}: This embedding is the one used as a baseline. In this scenario \(K=0\), \(k_1=0\) and \(k_2=0\), therefore \(\Phi(H_G) = \varphi(H_G^0)\). No abstract information is taken into consideration, hence, \(\Phi(H_G) = \varphi(G)\).

\textbf{Pyramidal embedding}: This embedding has been previously proposed in the literature~\cite{Mousavi2017,duttapyramidal}. It combines information of the abstract levels of the graph \ie~\(H_G^i\) not taking into account hierarchical information. Therefore, the hierarchical edges are discarded and no relation between levels is considered, \(K\geq1\), \(k_1=0\) and \(k_2=0\). We define \(\Phi_{\text{pyr}}(H_G) = [\varphi(H_G^0),\ldots,\varphi(H_G^K)]\). Note that each element corresponds to independent levels of the hierarchy without hierarchical edges.

\textbf{Generalized pyramidal embedding}: Following the previous idea, the information of the abstract levels of the graph \ie~\(H_G^i\) is combined. Now, hierarchical information is taken into account by embedding unions of levels \ie~\(H_G^{i_1} \cup H_G^{i_2}\) but discarding hierarchical edges (no clustering information is taken into account). In this scenario \(K\geq1\), \(k_1=0\) and \(k_2\geq1\), therefore, we define \(\Phi_{\text{gen\_pyr}}(H_G) = [\varphi(H_G^0), \ldots, \varphi(H_G^K), \varphi(H_G^0 \cup H_G^1), \ldots, \varphi(H_G^{K-1} \cup H_G^K), \ldots, \varphi(H_G^0 \cup \cdots \cup H_G^{k_2}), \ldots, \allowbreak \varphi(H_G^{K-k_2} \allowbreak \cup \cdots \cup H_G^K)]\).

\textbf{Hierarchical embedding}: This embedding is computed mixing different levels considering them as a single graph through the hierarchical edges, \(K \geq 1\), \(k_1 \geq 1\) and \(k_2=0\). The idea is to create an embedding able to codify both, graph and clustering information. Depending on the embedding, hierarchical edges can make use of special label to treat them differently. The Hierarchy embedding is defined as \(\Phi_{\text{hier}}(H_G) = [\varphi(H_G^0), \allowbreak \ldots, \allowbreak \varphi(H_G^K), \allowbreak \varphi(H_G^{0,1}),  \allowbreak \ldots, \allowbreak \varphi(H_G^{K-1,K}), \allowbreak \ldots, \allowbreak \varphi(H_G^{0,k_1}) \allowbreak, \ldots, \allowbreak \varphi(H_G^{K-k_1,K})]\). Note that each element corresponds to the subhierarchy compressed between the specified levels.

\textbf{Exhaustive embedding}: Finally, in order to take into consideration the whole hierarchy, we can make use of the whole embedding \(\Phi\) as defined in~\eq{eq1} where \(K \geq 1\), \(k_1, k_2 \geq 1\).

\fig{sfig:hierarchy_emb} shows the graphs taken into consideration when the hierarchical embeddings are computed.

\section{Stochastic graphlet embedding}
\label{sec:sge}
\begin{figure*}[t]
    \centering
    \includegraphics[width=\textwidth]{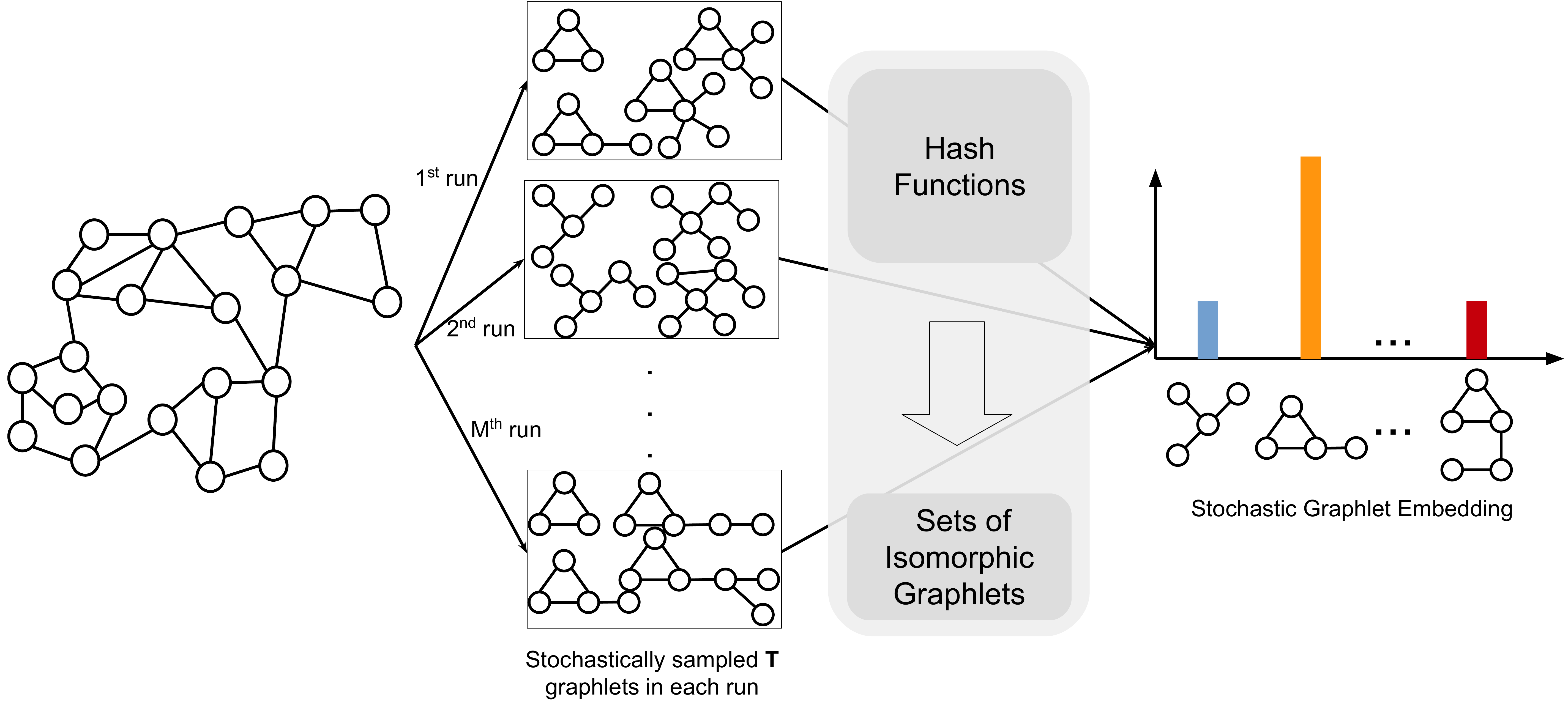}
    \caption{Overview of stochastic graphlet embedding (SGE). Given a graph $G$, the stochastic parsing algorithm is able to uniformly sample graphlets of increasing size. Controlled by two parameters $M$ (number of graphlets to be sampled) and $T$ (maximum size of graphlets in terms of number of edges), the method extracts in total $M\times T$ graphlets. These graphlets are encoded and partitioned into isomorphic graphlets using the set of hash functions with a low probability of collision. A distribution of different graphlets is obtained by counting the number of graphlets in each of these partitions. This procedure results in a vector space representation of the graph $G$ referred to as stochastic graphlet embedding.}
    \label{fig:sge}
\end{figure*}

The \emph{Stochastic Graphlet Embedding} (SGE) can be defined as a function $\varphi:\mathbb{G} \rightarrow \mathbb{R}^n$ that explicitly embeds a graph $G\in\mathbb{G}$ to a high dimensional vector space $\mathbb{R}^n$~\cite{Dutta2017a}. The entire procedure of SGE can be described in two stages (see~\fig{fig:sge}), where in the first step, the method samples graphlets from $G$ in a stochastic manner and in the second step, it counts the frequency of each isomorphic graphlet from the extracted ones in an approximated but near accurate manner. The entire procedure fetches a precise distribution of connected graphlets with increasing number of edges in $G$ with a controlled complexity, which fetches the relation among information represented as nodes and their complex interaction.

\subsection{Stochastic graphlets sampling}
\label{ssec:stoch-grphlet-sampl}
Considering a graph $G=(V,E,L_V,L_E)$, the goal of the graphlet extraction procedure is to obtain statistics of stochastic graphlets with increasing number of edges in $G$. The way of extracting graphlets is stochastic and it uniformly samples graphlets with boundlessly increasing number of edges without constraining their topology or structural properties such as maximum degree, maximum number of nodes, etc. Our graphlet sampling procedure, outlined in~\alg{alg:extractgrahletsdfs}, is recurrent and the number of recurrences is controlled by a parameter $M$ that indicates the number of distinct graphlets to be sampled (see~\lin{alg:extractgrahletsdfs:it1} of~\alg{alg:extractgrahletsdfs}). Also, each of these $M$ recurrent processes is regulated by another parameter $T$ that denotes the maximum number of iterations a single recurrent process should have (see~\lin{alg:extractgrahletsdfs:it2}). Since each of these iterations adds an edge to the presently constructing graphlet, $T$ indirectly specifies the maximum number of distinct edges each graphlet should contain. Considering $U_t$ and $A_t$ respectively as the aggregated sets of visited nodes and edges till iteration $t$, they are initialized at the beginning of each recurrent step as $A_0=\emptyset$ and $U_0=\lbrace u \rbrace$ with a randomly selected node $u$ which is uniformly sampled from $V$ (see~\lin{alg:extractgrahletsdfs:init}). Thereafter, at $t^\text{th}$ iteration (with $t\ge1$), the sampling procedure randomly selects an edge $(u,v)\in E \backslash A_{t-1}$ that is connected from any node $u\in U_{t-1}$ (see~\lin{alg:extractgrahletsdfs:ransel3}). Accordingly, the process updates $U_t \leftarrow U_{t-1} \cup \lbrace v \rbrace$ and $A_{t} \leftarrow A_{t-1} \cup \lbrace (u,v) \rbrace$ (see~\lin{alg:extractgrahletsdfs:updateVN}). All these processes within a recurrent step are repeated $T$ times to sample a graphlet with maximum $T$ edges. $M$ is set to relatively large values in order to make the graphlet generation statistically meaningful. Theoretically, the values of $M$ are guided by the theorem of \emph{sample complexity}~\cite{Weissman2003}, which is widely studied and used in the Bioinformatics domain~\cite{Przulj2007,Shervashidze2009}. However, the discussion and proof of that is out of scope of the current paper. Intuitively, the graphlet sampling procedure explained in this section follows a random walk process with restart that efficiently parses $G$ and extracts the desired number of connected graphlets with an increasing number of edges. This algorithm allows to sample connected graphlets from a given graph but avoids expensive way of extracting them in an exact manner. Here the hypothesis is that if a sufficient number of graphlets are sampled, then the empirical distribution will be close to the actual distribution of graphlets in the graph. Furthermore, it is important to note that from the above process, one can extract, in total, $M \times T$ graphlets each with number of edges varying from $1$ to $T$.

\begin{algorithm}
\caption{\textsc{Stochastic-Graphlet-Parsing}($G$): Create a set of graphlets $\mathbb{S}$ by traversing $G$.}
\label{alg:extractgrahletsdfs}
\begin{algorithmic}[1]
\REQUIRE $G=(V,E)$, $\mathit{M}$, $\mathit{T}$
\ENSURE $\mathbb{S}$
\STATE $\mathbb{S}\leftarrow \emptyset$
\FOR {$i=1$ to $\mathit{M}$} \label{alg:extractgrahletsdfs:it1}
    \STATE $u \leftarrow \textsc{SelectRandomNode}(V)$ \label{alg:extractgrahletsdfs:ransel1}
    \STATE $\mathit{U_0}\leftarrow u$, $\mathit{A_0} \gets \emptyset$ \label{alg:extractgrahletsdfs:init}
    \FOR {$t=1$ to $\mathit{T}$} \label{alg:extractgrahletsdfs:it2}
        \STATE $u \leftarrow \textsc{SelectRandomNode}(\mathit{U_{t-1}})$ \label{alg:extractgrahletsdfs:ransel2}
        \STATE $v \leftarrow \textsc{SelectRandomNode}(V): (u,v)\in E\setminus\mathit{A_{t-1}}$ \label{alg:extractgrahletsdfs:ransel3}
        \STATE $U_t \leftarrow U_{t-1} \cup \{v\}$ , \ \ \ $A_t \leftarrow A_{t-1} \cup \{(u,v)\}$ \label{alg:extractgrahletsdfs:updateVN}
        \STATE $\mathbb{S} \leftarrow \mathbb{S} \cup\lbrace( U_t, A_t)\rbrace$
    \ENDFOR
\ENDFOR
\end{algorithmic}
\end{algorithm}

\subsection{Hashed graphlets distribution}
\label{ssec:hash-grphlet-stats}
For obtaining a distribution of the extracted graphlets from $G$, it is needed to identify sets of isomorphic graphlets from the sampled ones and then count cardinality of each isomorphic set. A trivial way of doing that certainly involves checking the graph isomorphism for all possible pairs of graphlets for detecting possible partitions that might exist among them. Nevertheless, graph isomorphism is a GI-complete problem~\cite{Mehlhorn1984} for general graphs, so the previously mentioned scheme is extremely costly as the method samples huge number of graphlets with many edges. An alternative, efficient and approximate way of partitioning isomorphic graphlets is \emph{graph hashing}. A graph hash function can be defined as a mapping $h:\mathbb{G} \rightarrow \mathbb{R}^m$ that maps a graph into a hash code (a sequence of real numbers) based on the local as well as holistic topological characteristic of graphs. An ideal graph hash function should map two isomorphic graphs to the same hash code as well as two non-isomorphic graphs to two different hash codes. While it is easy to design hash functions satisfying the condition that two isomorphic graphs should have the same hash code, it is extremely difficult to find hash function that ensures different hash codes for every pair of non-isomorphic graphs. An alternative is to design graph hash functions with low \emph{collision probability}, \ie, mapping any two non-isomorphic graphs to the same hash code with a very low probability. For obtaining a distribution of graphlets, the main aim of graph hashing is to assign extracted graphlets from $G$ to corresponding subsets of isomorphic graphlets (a.k.a. partition index or histogram bins) in order to count and quantify their distributions. The proposed mechanism for obtaining the distribution of uniformly sampled graphlets, outlined in~\alg{alg:graphlethistogram}, maintains a global hash table $\mathbf{H}$, whose single entry corresponds to a hash code of a graphlet $g$ produced by the graph hash function. $\mathbf{H}$ grows incrementally as the algorithm confronts new graph hash codes and maintains all the unique hash codes encountered by the system. It is to be noted that the position of each unique hash code is kept fixed, because each position corresponds to a partition index or histogram bin. Now to allocate a given graphlet $g$ to its corresponding histogram bin, its hash code $h(g)$ is mapped to the index of the hash table $\mathbf{H}$, whose corresponding graph hash code gives a hit with $h(g)$ (see~\lin{alg:graphlethistogram:idx}). If $h(g)$ does not exist in $\mathbf{H}$ at some instance, it is considered as a new hash code (and hence $g$ as a new graphlet) encountered by the system and appended $h(g)$ at the end of $\mathbf{H}$ (see~\lin{alg:graphlethistogram:newhashcode}).

\begin{algorithm}
\caption{\textsc{Hashed-Graphlets-Statistics}($G$): Create a histogram $\mathbf{h}$ of graphlet distribution for a graph $G$.}
\label{alg:graphlethistogram}
\begin{algorithmic}[1]
\REQUIRE $G$, $\mathbf{H}$\label{alg:graphlethistogram:hashtable}
\ENSURE $\mathbf{h}$
\STATE $\mathbb{S} \leftarrow \textsc{Stochastic-Graphlet-Parsing}(G)$
\STATE $\mathbf{h}_i\leftarrow 0$, $i=1,\dots,|\mathbb{S}|$
\FORALL {$g \in \mathbb{S}$}
    \STATE $h(g) \leftarrow \textsc{HashFunction}(g)$\label{alg:graphlethistogram:hashcode}
    \IF {$h(g)\notin\mathbf{H}$}
    \STATE $\mathbf{H}\leftarrow \mathbf{H} \cup \{h(g)\} $\label{alg:graphlethistogram:newhashcode}
    \ENDIF
    \STATE $i \leftarrow \textsc{GetIndex-In-HashTable}(h(g))$\label{alg:graphlethistogram:idx}
    \STATE $\mathbf{h}_i \leftarrow \mathbf{h}_i + 1$\label{alg:graphlethistogram:hist}
\ENDFOR
\end{algorithmic}
\end{algorithm}

Designing hash functions that yield identical hash codes for two isomorphic graphlets is quite simple, whereas, prototyping those providing two distinct hash codes for two non-isomorphic graphs is very challenging. The chance of mapping two non-isomorphic subgraphs to the same hash code is termed as \emph{probability of collision}. Indicating $H_0$ as the set of all pairs of non-isomorphic graphs, the probability of collision can be expressed as the following energy function:
\begin{equation}
    E(f) = P((g,g') \in H_0 \quad | \quad h(g) = h(g'))
    \label{eqn:energy_collis}
\end{equation}
So, in terms of collision probability, the hash functions that produce comparatively lower $E(f)$ values in~\eq{eqn:energy_collis} are considered to be more reliable for checking the graph isomorphism. It has been studied that sorted \emph{degree of nodes} has $0$ collision probability for all graphs with number of edges less or equal to $4$~\cite{Dutta2017a}. Moreover, it is also a well known fact that two graphs with the same \emph{betweenness centrality} (sorted) would indeed be isomorphic with high probability~\cite{Comellas2008,Newman2005}. For example, sorted \emph{betweenness centrality} has collision probabilities equal to $3.2e^{-4}$, $1.9e^{-4}$, $1.1e^{-4}$ respectively for graphlets with $7$, $8$ and $9$ edges. Interested readers are requested to see~\cite{Dutta2017a} for further discussions and analysis on various graph hash functions and corresponding elaboration on probability of collision. Considering the above facts, in this work, we consider sorted \emph{degree of nodes} for graphlets with $t\leq 4$ and the \emph{betweenness centrality} for graphlets with $t\ge 5$.
\begin{equation}
\text{Hash function}=
    \begin{cases}
        \text{ degree of nodes},& \text{if } t\leq 4\\
        \text{ betweenness centrality},& \text{otherwise}
    \end{cases}
\end{equation}
It should be observed that the distribution of sampled graphlets obtained the way mentioned until now, only considers the topological structure of a graph, and ignores the node and edge attributes. However, it is worth mentioning that the stochastic graphlet embedding permits to consider a small set of nodes and edge attributes by creating respective signatures and then appending it to the hash code encoding the topology of the graphlet. In this work, if needed, we first discretize the existing continuous attributes using a combination of clustering algorithm such as \emph{k-means} and pooling technique. Later, the sorted discrete node and edge labels are used as the attribute signatures and combined with the hash code.

\subsection{Hierarchical stochastic graphlet embedding}
\label{ssec:hsge}
In this work, we propose to combine the properties of the proposed \emph{Stochastic Graphlet Embedding} with the \emph{Hierarchical Embedding} introduced in the previous Section.

On one hand, SGE provides statistical information about local structures varying the number of edges involved. Therefore, it provides fine-grained insights of the graph which cannot deal with too noisy data. The use of abstractions provided by the graph hierarchy increases the receptive field of each graphlet moving to coarser information that is able to provide insights of the global graph information. Moreover, the use of hierarchical edges during the computation allows to combine information at some levels, \emph{i.e.} combining different levels of detail (see~\eq{eq1}). For now on, we will denote this embedding as \emph{Hierarchical Stochastic Graphlet Embedding}.

\section{Computational complexity}
\label{sec:compl}

This Section is devoted to study the computational complexity of the proposed approach given a graph \(G=(V,E,L_V,L_E)\) where \(|V|=n\) and \(|E|=m\).

\subsection{Hierarchical embedding complexity}
\label{ssec:compl_hierarchical}

Graph clustering algorithms are usually high computational complexity techniques. As it has been stated in~\sect{ssec:hembed}, the \emph{Girvan-Newman} algorithm has been chosen as a graph clustering technique. The Girvan-Newman algorithm is based on the betweenness centrality of the edges which has a time complexity of \(\mathcal{O}(n \cdot m)\) for unweighted graphs and \(\mathcal{O}(n \cdot m + n\cdot (n+m) \log(n))\) for weighted graphs. Hence, the \emph{Girvan-Newman} algorithm, which has to remove all the edges, can be computed in \(\mathcal{O}(n \cdot m^2)\) for unweighted graphs and \(\mathcal{O}(n \cdot m^2 + n\cdot m \cdot (n+m) \log(n))\) for weighted graphs.

Assuming an embedding function \(\varphi\) which has a complexity of \(\mathcal{O}(N)\) and assuming that the hierarchical graph construction has a complexity of \(C_1\), then, if we assume \(L\) levels, the proposed configurations would become a complexity \(\mathcal{O}(C_1 + L\cdot N)\) in the case of the pyramid and \(\mathcal{O}(C_1 + L^2\cdot N)\) for the hierarchy and the exhaustive embeddings.

\subsection{Stochastic graphlet embedding complexity}
\label{ssec:compl_sge}

The computational complexity of~\alg{alg:extractgrahletsdfs} is \(\mathcal{O}(M \cdot T)\) where \(M\) is the number of graphlets to be sampled and \(T\) is the maximum size of graphlets in terms of the number of edges. Assuming a hash function with a complexity of \(\mathcal{O}(C_2)\),~\alg{alg:graphlethistogram} has a time complexity of $\mathcal{O}(M \cdot T \cdot C_2)$ for computing the stochastic graphlet embedding. Here it is worth mentioning that ``degree of nodes'' and ``betweeness centrality'', respectively have the time complexity of $\mathcal{O}(n)$ and $\mathcal{O}(n \cdot m)$. From the above explanation, it is clear that the complexity of these two algorithms do not depend on the size of the input graph $G$, but only on the parameters $M$, $T$ and the hash functions used.

\section{Experimental validation}
\label{sec:expt}
This section presents the experimental results obtained by our proposed Hierarchical Stochastic Graphlet Embedding method. The main aim of this experimental study is to validate the proposed graph embedding technique for the graph classification task, which demands robust embedding technique for mapping a graph into a vector space. For experimentation, we have considered many different widely used graph datasets with varied characteristics. All these graphs come from real data generated in the fields of Biology, Chemistry, Graphics and Handwriting recognition. The MATLAB code of our experiment is available at \url{https://github.com/priba/hierarchicalSGE}.

\subsection{Experiments on molecular graph datasets}
\label{sssec:molecular}
The first set of experiments is conducted on various benchmarks of molecular graphs. Below, we provide a brief description of them followed by the experimental setup, results and discussions.

\subsubsection{Dataset description}
Several bioinformatics datasets have been used: \emph{MUTAG}, \emph{PTC}, \emph{PROTEINS}, \emph{NCI1}, \emph{NCI109}, \emph{D}\&\emph{D} and \emph{MAO}. These datasets have been widely used as benchmark in the literature. The \emph{MUTAG} dataset contains graph representations of 188 chemical compounds which are either mutagenic aromatic or heteroromatic nitro compounds where nodes can have 7 discrete labels. The \emph{PTC} or Predictive Toxicology Challenge dataset consists of 344 chemical compounds known to cause or not cause cancer in rats and mice. It has 19 discrete node labels. The \emph{PROTEINS} dataset contains relations between secondary structure elements (SSEs) represented by nodes and neighborhood in the amino-acid sequence or in 3D space by edges. It has 3 discrete labels \viz~\emph{helix}, \emph{sheet} or \emph{turn}. The \emph{NCI1} and \emph{NCI109} come from the National Cancer Institute (NCI) and are two balanced subsets of chemical compounds screened for their ability to suppress or inhibit the growth of a panel of human tumor cell lines, having 37 and 38 discrete node labels respectively. The \emph{D}\&\emph{D} dataset consists of enzymes and non-enzymes proteins structures, in which their nodes are amino acids. The \emph{MAO} database, taken from GREYC Chemistry graph dataset collection, is composed of 68 graphs representing molecules that either inhibit or not the monoamine oxidase, which is an antidepressant drug. Some more details on the proposed bioinformatics datasets are provided in~\tab{t:bioinformatics}.

\begin{table}[htb]
\centering
\caption{Details of the molecular graph datasets.}
\label{t:bioinformatics}
\resizebox{\columnwidth}{!}{
\begin{tabular}{l c c c c c c}
\toprule
\textbf{Datasets} & \textbf{\# Graphs} & \textbf{\# Classes} & \textbf{Avg. \(|V|\)} & \textbf{Avg. \(|E|\)} & \textbf{Node labels}\\
\midrule
\textbf{MUTAG} & 188 & 2 (125 vs. 63) & 17.9 & 39.6 & 7  \\
\textbf{PTC} & 344 & 2 (192 vs. 152) & 25.6 & 51.9 & 19  \\
\textbf{PROTEINS} & 1113 & 2 (663 vs. 450) & 39.1 & 145.63 & 3 \\
\textbf{NCI1} & 4110 & 2 (2057 vs. 2053) & 29.9 & 64.6 & 37 \\
\textbf{NCI109} & 4127 & 2 (2079 vs. 2048) & 29.7 & 64.3 & 38 \\
\textbf{D\&D} & 1178 & 2 (691 vs 487) & 284.3 & 1431.3 & 82 \\
\textbf{MAO} & 68 & 2 (30 vs. 38) & 18.4 & 19.6 & 3 \\
\bottomrule
\end{tabular}}
\end{table}

\subsubsection{Experimental setup}
We have performed two different experiments: the first one does not use the attribute information encoded in the nodes and edges of the graphs, whereas the second experiment does use the available node and edge features. For evaluating the performance of the proposed embedding technique, we have used a C-SVM solver~\cite{chang2011libsvm} as a classifier. Since the datasets considered in this set of experiments do not contain predefined train and test sets, we have used a $10$-fold cross validation scheme to obtain accuracies and have reported the mean accuracies respectively in~\tab{t:unlabeledBio} and~\tab{t:labeledBio} for unlabeled and labeled datasets. We follow a classical graph classification pipeline where, in the first stage, graph embedding is computed by our proposed scheme, whereas in the second step, embedded graphs are classified using a previously trained classifier.

\subsubsection{Results and discussion}
In~\tab{t:unlabeledBio}, we present the experimental results obtained by our proposed hierarchical embedding techniques together with other existing works on the unlabeled datasets. The previously mentioned three configurations of our hierarchical embedding are respectively denoted as: pyramidal, hierarchical and exhaustive. For unlabeled datasets, we have considered $10$ different state-of-the-art methods: (1) random walk kernel (RW)~\cite{Gartner2003}, (2) shortest path kernel (SP)~\cite{Borgwardt2005}, (3) graphlet kernel (GK)~\cite{Shervashidze2009}, (4) Weisfeiler-Lehman kernel (WL)~\cite{Shervashidze2011}, (5) deep graph kernel (DGK)~\cite{Yanardag2015}, (6) multiscale Laplacian graph kernel (MLK)~\cite{Kondor2016}, (7) diffusion CNNs (DCNN)~\cite{Atwood2016}, (8) strong graph spectrums (SGS)~\cite{Kondor2008}, (9) family of graph spectral distances (F\_GSD)\allowbreak~\cite{Verma2017}, and (10) stochastic graphlet embedding (SGE)~\cite{Dutta2017a}.

From the quantitative results shown in~\tab{t:unlabeledBio}, it should be observed that for most datasets, the highest accuracy is achieved by one of the hierarchical configurations proposed by us, which sets a new state-of-the-art results on all the datasets considered. Particularly, the best accuracies are obtained either by the pyramidal or the exhaustive configurations, which indicates the importance of considering hierarchical information for the graph embedding problem. As expected, the proposed hierarchical embeddings have achieved better performance than the SGE which is regarded as the baseline of our proposal. It should be observed that with this experimental setting, particularly the hierarchical configuration has performed quite poorly compared to the other two configurations. This fact might suggest that only hierarchical edges together with the connecting levels do not contain sufficient information for a robust graph representation. Information captured in the multiscale graphs thought to play a vital role for graph embedding, which is proved by the excellent performance obtained with the pyramidal and exhaustive configurations.

\begin{table*}[htb]
\centering
\caption{Classification accuracies on \emph{unlabeled} molecular graph datasets. In the table, RW corresponds to the random walk kernel~\cite{Gartner2003}, SP stands for the shortest path kernel~\cite{Borgwardt2005}, GK denotes the graphlet kernel~\cite{Shervashidze2009}, WL indicates the Weisfeiler-Lehman kernel~\cite{Shervashidze2011}, DGK corresponds to the deep graph kernel~\cite{Yanardag2015}, MLK stands for the multiscale Laplacian graph kernel~\cite{Kondor2016}, DCNN indicates the diffusion CNNs~\cite{Atwood2016}, SGS denotes the strong graph spectrums~\cite{Kondor2008}, F\_GSD stands for the family of graph spectral distances~\cite{Verma2017}, SGE corresponds to the stochastic graphlet embedding~\cite{Dutta2017a}, and HSGE indicates the hierarchical graph embeddings proposed by us. The best results obtained on a dataset is indicated by bold face.}
\label{t:unlabeledBio}
\resizebox{\textwidth}{!}{
\begin{tabular}{l c c c c c c c}
\toprule
\textbf{Methods} & \textbf{MUTAG} & \textbf{PTC} & \textbf{PROTEINS} & \textbf{NCI1} & \textbf{NCI109} & \textbf{D\&D} & \textbf{MAO} \\
\midrule
\textbf{RW}~\cite{Gartner2003} & $83.50$ & $55.52$ & $68.46$ & $44.84$ & $59.80$ & $-$ & $83.52$ \\
\textbf{SP}~\cite{Borgwardt2005} & $87.23$ & $58.72$ & $72.14$ & $68.15$ & $68.30$ & $-$ & $90.35$ \\
\textbf{GK}~\cite{Shervashidze2009} & $84.04$ & $60.17$ & $71.78$ & $62.07$ & $62.04$ & $75.05$ & $80.88$ \\
\textbf{WL}~\cite{Shervashidze2011} & $87.28$ & $55.61$ & $70.06$ & $77.23$ & $78.43$ & $73.76$ & $89.79$ \\
\textbf{DGK}~\cite{Yanardag2015} & $86.17$ & $59.88$ & $71.69$ & $64.40$ & $67.14$ & $72.75$ & $87.76$ \\
\textbf{MLK}~\cite{Kondor2016} & $87.23$ & $62.20$ & $71.35$ & $77.57$ & $75.91$ & $77.02$ & $91.17$ \\
\textbf{DCNN}~\cite{Atwood2016} & $66.51$ & $55.79$ & $65.22$ & $63.10$ & $60.67$ & OMR & $76.10$ \\
\textbf{SGS}~\cite{Kondor2008} & $88.61$ & $-$ & $-$ & $62.72$ & $62.62$ & $-$ & $-$ \\
\textbf{F\_GSD}~\cite{Verma2017} & $92.12$ & $62.80$ & $73.42$ & $79.80$ & $78.84$ & $77.10$ & $95.58$ \\
\textbf{SGE}~\cite{Dutta2017a} & $91.11$ & $63.53$ & $71.89$ & $83.23$ & $82.92$ & $74.92$ & $95.71$ \\
\textbf{HSGE (pyr.)} & $91.11$ & $65.29$ & $75.32$ & $\mathbf{85.24}$ & $\mathbf{83.24}$ & $78.73$ & $\mathbf{100.00}$ \\
\textbf{HSGE (gen. pyr.)} & $92.22$ & $67.94$ & $75.50$ & $83.36$ & $81.73$ & $\mathbf{79.32}$ & $\mathbf{100.00}$ \\
\textbf{HSGE (hier.)} & $\mathbf{93.33}$ & $67.06$ & $\mathbf{76.31}$ & $82.85$ & $81.33$ & $72.03$ & $\mathbf{100.00}$ \\
\textbf{HSGE (exhaus.)} & $92.22$ & $\mathbf{70.88}$ & $76.58$ & $83.84$ & $82.12$ & $73.90$ & $\mathbf{100.00}$ \\
\bottomrule
\end{tabular}}
\end{table*}

In~\tab{t:labeledBio}, we demonstrate the results acquired by three different configurations of our proposed hierarchical embedding on the labeled graph datasets. For comparing with other state-of-the-art methods, we have considered two additional techniques: (1) PATCHY-SAN (PSCN)~\cite{Niepert2016} and (2) graphlet spectrum (GS)~\cite{Kondor2009}. Some of the previously considered state-of-the-art techniques do not work with labeled graphs, so they have not been evaluated in this experimentation.

The results presented in~\tab{t:labeledBio} show that, except on the MUTAG dataset, our proposed hierarchical embedding techniques have achieved the best performances on all the other datasets. This demonstrates the usefulness of considering the hierarchical information for embedding graphs to a vector space. Contrary to the previous experiments on unlabeled datasets, in this case, the hierarchical configuration has performed reasonably better. This fact shows that on labeled graphs, the hierarchical edges together with the connecting levels might provide important structural information. Also, it is important to note that the level information also performed consistently on all the datasets.

\begin{table*}[htb]
\centering
\caption{Classification accuracy on \emph{labeled} molecular graph datasets. In the table, MLK stands for the Multiscale Laplacian Graph kernel~\cite{Kondor2016}, DCNN indicates the Diffusion CNNs~\cite{Atwood2016}, PSCN corresponds to the PATCHY-SAN~\cite{Niepert2016}, GS denotes the Graphlet Spectrum (GS)~\cite{Kondor2009}, SGE corresponds to the Stochastic Graphlet Embedding (SGE)~\cite{Dutta2017a}, and HSGE indicates the hierarchical graph embeddings proposed by us. The best results obtained on a dataset is specified by bold face.}
\label{t:labeledBio}
\resizebox{\textwidth}{!}{
\begin{tabular}{l c c c c c c c}
\toprule
\textbf{Methods} & \textbf{MUTAG} & \textbf{PTC} & \textbf{PROTEINS} & \textbf{NCI1} & \textbf{NCI109} & \textbf{D\&D} & \textbf{MAO} \\
\midrule
\textbf{MLK}~\cite{Kondor2016} & $87.94$ & $63.26$ & $-$ & $81.75$ & $-$ & $78.18$ & $88.29$ \\
\textbf{DCNN}~\cite{Atwood2016} & $66.98$ & $56.60$ & $-$ & $62.61$ & $-$ & OMR & $75.14$ \\
\textbf{PSCN}~\cite{Niepert2016} & $92.63$ & $62.90$ & $-$ & $78.59$ & $-$ & $77.12$ & $-$ \\
\textbf{GS}~\cite{Kondor2009} & $88.11$ & $-$ & $-$ & $65.00$ & $-$ & $-$ & $-$ \\
\textbf{SGE}~\cite{Dutta2017a} & $88.33$ & $57.94$ & $74.05$ & $83.44$ & $81.89$ & $77.37$ & $94.29$ \\
\textbf{HSGE (pyr.)} & $91.11$ & $62.06$ & $75.68$ & $\mathbf{84.79}$ & $82.03$ & $77.46$ & $94.29$ \\
\textbf{HSGE (gen. pyr.)} & $\mathbf{92.78}$ & $65.59$ & $\mathbf{76.58}$ & $81.31$ & $80.24$ & $79.66$ & $\mathbf{97.14}$ \\
\textbf{HSGE (hier.)} & $91.11$ & $\mathbf{67.35}$ & $75.77$ & $82.50$ & $82.88$ & $79.32$ & $94.29$ \\
\textbf{HSGE (exhaust.)} & $91.67$ & $66.18$ & $76.04$ & $84.42$ & $\mathbf{84.42}  $ & $\mathbf{80.25}$ & $\mathbf{97.14}$ \\
\bottomrule
\end{tabular}}
\end{table*}

\subsection{Experiments on AIDS, GREC, COIL-DEL and HistoGraph datasets}
While the datasets considered in the previous set of experiments were mostly molecular in nature, the set of experiments to be discussed in this section consider graphs from various fields, such as, Biology, Computer Vision, Graphics Recognition and Handwriting Recognition. Underneath, we give a brief description of the datasets considered followed by the experimental setup, results and discussions.

\subsubsection{Dataset description}
In this experiment, we consider four different datasets; three of them \viz~\emph{AIDS}, \emph{GREC} and \emph{COIL-DEL} are taken from the IAM graph database repository\footnote{Available at \url{http://www.fki.inf.unibe.ch/databases/iam-graph-database}}~\cite{Riesen2008}. The first one, \viz, the AIDS database consists of $2000$ graphs representing molecular compounds which are constructed from the AIDS Antiviral Screen Database of Active Compounds\footnote{See at \url{http://dtp.nci.nih.gov/docs/aids/aids_data.html}}. This dataset consists of two classes, \viz, active ($400$ elements) and inactive ($1600$ elements), which respectively represent molecules with possible activity against HIV. The GREC dataset consists of $1100$ graphs representing $22$ different classes (characterizing architectural and electronic symbols) with $50$ instances per class; these instances have different noise levels. The COIL-DEL database includes $3900$ graphs belonging to $100$ different classes with $39$ instances per class; each instance has a different rotation angle. The HistoGraph dataset\footnote{Available at \url{http://www.histograph.ch}}\cite{Stauffer2016} consists of graphs representing words from the communicating letters written by the first US president, George Washington. It consists of $293$ graphs generated from $30$ distinct words. Therefore, given a word, the task of the classifier is to predict its class which should be among the $30$ words. Nodes are only labeled with their position in the image. Furthermore, this dataset used $6$ different graph representation paradigms for delineating a single word into a graph, which results in $6$ different subsets of graphs. The entire dataset is divided into $90$, $60$ and $143$ graphs respectively for train, validation and test purposes.

\begin{table}[htb]
\centering
\caption{Details of the AIDS, GREC, COIL-DEL and HistoGraph datasets.}
\label{t:patternreco}
\resizebox{\columnwidth}{!}{
\begin{tabular}{l l c c c c c}
\toprule
\textbf{Datasets} & \textbf{Subsets} & \textbf{\# Graphs} & \textbf{\# Classes} & \textbf{Avg. \(|V|\)} & \textbf{Avg. \(|E|\)} & \textbf{Node labels} \\
\midrule
\textbf{AIDS}& $-$ & $2000$ $(250, 250, 1500)$ & $2$  & $15.7$ & $16.2$ & Chemical symbol \\
\textbf{GREC}& $-$ & $1100$ $(286, 286, 528)$ & $22$ ($50$ each) & $11.5$ & $11.9$ & Type, (x,y) position \\
\textbf{COIL-DEL}& $-$ & $3900$ $(2400, 500, 1000)$ & $100$ & $21.5$ & $54.2$ & (x,y) position\\
\multirow{6}{*}{\textbf{HistoGraph}} & \textbf{Keypoint} & \multirow{6}{*}{$293$ $(90,60,143)$} & \multirow{6}{*}{$30$} & $101.8$ & $94.8$ & \multirow{6}{*}{(x,y) position} \\
& \textbf{Grid-NNA} &  & & $56.4$ & $81.4$ & \\
& \textbf{Grid-MST} &  & & $66.1$ & $64.4$ & \\
& \textbf{Grid-DEL} &  & & $74.1$ & $205.1$ & \\
& \textbf{Projection} &  & & $63.1$ & $58.8$ & \\
& \textbf{Split} &  & & $73.3$ & $69.8$ &  \\
\bottomrule
\end{tabular}}
\end{table}

\subsubsection{Experimental setup}
In this case as well, we have employed a C-SVM solver~\cite{chang2011libsvm} as a classifier. Since the datasets used in this set of experiments contain well defined train and test sets, we have reported the obtained accuracies on the test set of the respective datasets in~\tab{t:pr}.

\subsubsection{Results and discussion}
Similar to the experimental results obtained in the previous section, in this set of experiments as well, our proposed hierarchical embeddings have achieved the best results on most datasets. In this set of experiments, the leading scores are mostly obtained by the exhaustive configuration, which shows the effectiveness of combining multiscale structural information together with the hierarchical connections. For some datasets, our hierarchical embedding does not achieve the best results, but it has performed very competitively. This also proves the robustness of the hierarchical graph representation.

\begin{table*}[htb]
\centering
\caption{Results obtained on the AIDS, GREC, COIL-DEL and HistoGraph datasets. In the table, RW corresponds to the Random Walk kernel~\cite{Gartner2003}, DE stands for the dissimilarity embedding~\cite{Bunke2010}, NAS indicates the node attribute statistics~\cite{Gibert2012}, GED denotes to the approximated graph edit distance~\cite{Riesen2009}, SGE corresponds to the Stochastic Graphlet Embedding (SGE)~\cite{Dutta2017a}, and HSGE indicates our hierarchical graph embeddings. The best results obtained on a dataset is indicated by bold face.}
\label{t:pr}
\resizebox{\textwidth}{!}{%
\begin{tabular}{l c c c c c c c c c}
\toprule
\multirow{2}{*}{\textbf{Methods}} & \multirow{2}{*}{\textbf{AIDS}} & \multirow{2}{*}{\textbf{GREC}} & \multirow{2}{*}{\textbf{COIL-DEL}} & \multicolumn{6}{c}{\textbf{HistoGraph}} \\
  &  &  &  & \textbf{Keypoint} & \textbf{Grid-NNA} & \textbf{Grid-MST} & \textbf{Grid-DEL} & \textbf{Projection} & \textbf{Split} \\
\midrule
\textbf{RW}~\cite{Gartner2003} & $98.50$ & $96.20$ & $94.20$ & $-$ & $-$ & $-$ & $-$ & $-$ & $-$ \\
\textbf{DE}~\cite{Bunke2010} & $98.10$ & $95.10$ & $96.80$ & $-$ & $-$ & $-$ & $-$ & $-$ & $-$ \\
\textbf{NAS}~\cite{Gibert2012} & $98.30$ & $99.20$ & $98.10$ & $-$ & $-$ & $-$ & $-$ & $-$ & $-$ \\
\textbf{GED}~\cite{Riesen2009} & $-$ & $-$ & $-$ & $77.62$ & $65.03$ & $74.13$ & $62.94$ & $\mathbf{81.82}$ & $80.42$ \\
\textbf{SGE}~\cite{Dutta2017a} & $98.67$ & $\mathbf{99.62}$ & $98.14$ & $79.02$ & $\mathbf{72.73}$ & $77.62$ & $74.83$ & $79.72$ & $81.12$ \\
\textbf{HSGE (pyr.)} & $98.87$ & $99.43$ & $98.79$ & $79.02$ & $\mathbf{72.73}$ & $77.62$ & $74.83$ & $79.72$ & $81.12$ \\
\textbf{HSGE (gen. pyr.)} & $98.35$ & $99.43$ & $98.37$ & $77.62$ & $72.03$ & $77.62$ & $74.13$ & $79.72$ & $81.45$ \\
\textbf{HSGE (hier.)} & $98.33$ & $99.05$ & $\mathbf{98.99}$ & $79.02$ & $70.63$ & $76.22$ & $\mathbf{75.52}$ & $80.42$ & $80.42$ \\
\textbf{HSGE (exhaust.)} & $\mathbf{99.00}$ & $99.43$ & $98.86$ & $\mathbf{79.72}$ & $72.03$ & $\mathbf{78.32}$ & $74.83$ & $\mathbf{81.82}$ & $\mathbf{81.82}$ \\
\bottomrule
\end{tabular}
}
\end{table*}

\begin{figure*}[!htb]
    \centering
    \subfloat[]{\includegraphics[width=0.44\textwidth]{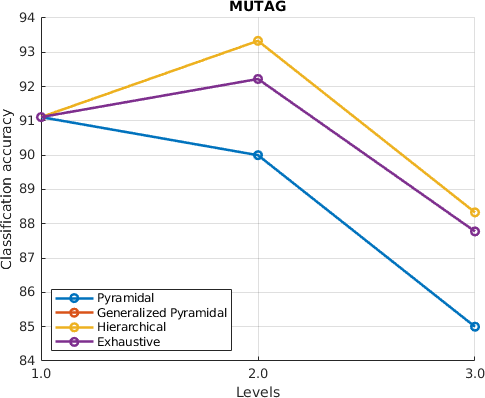}}
    \hspace{0.02\textwidth}
    \subfloat[]{\includegraphics[width=0.44\textwidth]{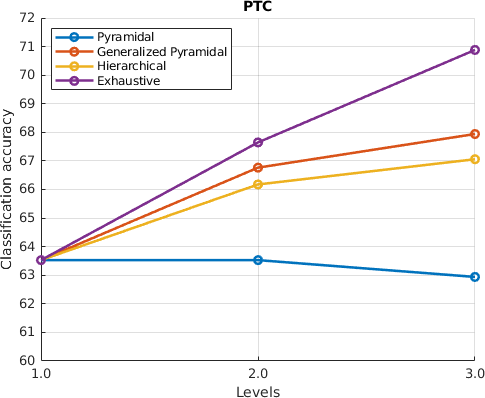}}
    \hspace{0.02\textwidth}
    \subfloat[]{\includegraphics[width=0.44\textwidth]{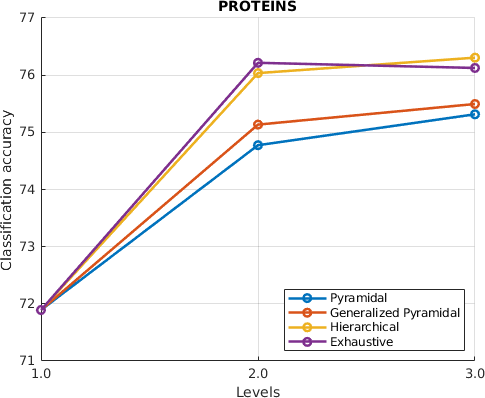}}
    \hspace{0.02\textwidth}
    \subfloat[]{\includegraphics[width=0.44\textwidth]{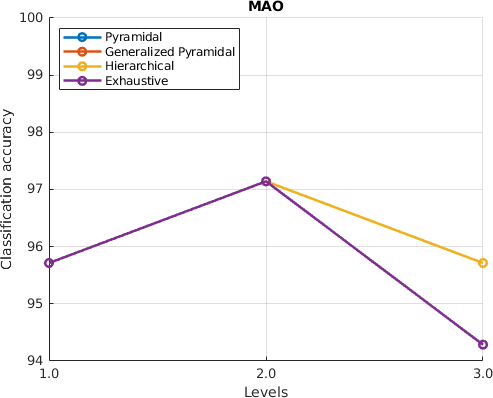}}
    \caption{Plots showing classification accuracies by varying the levels of pyramidal graph construction on different datasets.}
    \label{fig:plots_levels}
\end{figure*}

\begin{figure*}[!htb]
    \centering
    \subfloat[]{\includegraphics[width=0.44\textwidth]{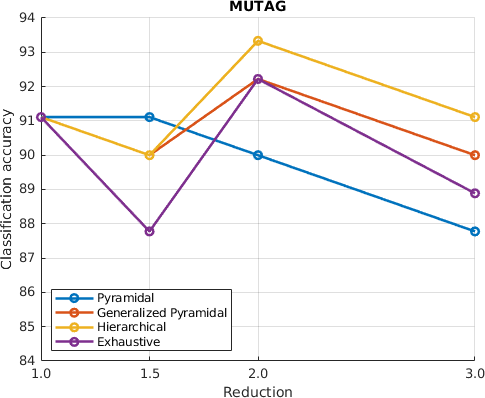}}
    \hspace{0.02\textwidth}
    \subfloat[]{\includegraphics[width=0.44\textwidth]{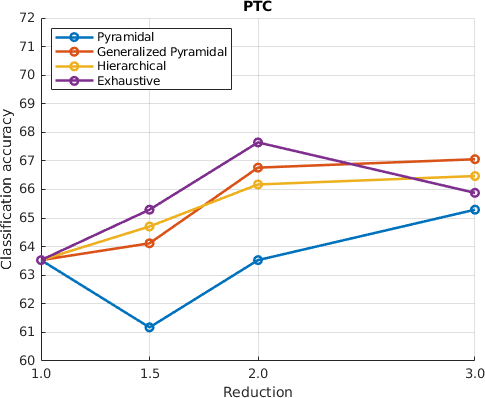}}
    \hspace{0.02\textwidth}
    \subfloat[]{\includegraphics[width=0.44\textwidth]{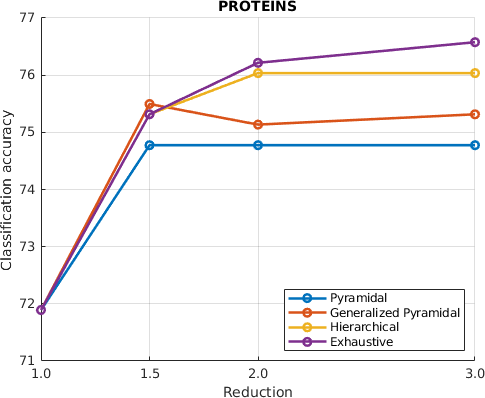}}
    \hspace{0.02\textwidth}
    \subfloat[]{\includegraphics[width=0.44\textwidth]{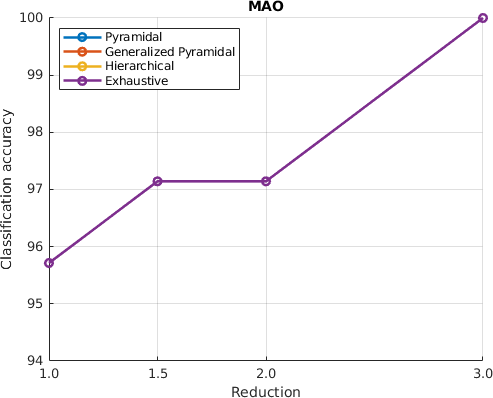}}
    \caption{Plots showing classification accuracies by varying the reduction ratio of pyramidal graph construction on different datasets.}
    \label{fig:plots_reduction}
\end{figure*}

\subsection{Discussion on the parameters involved in the algorithm}
\begin{figure}
\centering
\includegraphics[width=0.4\textwidth]{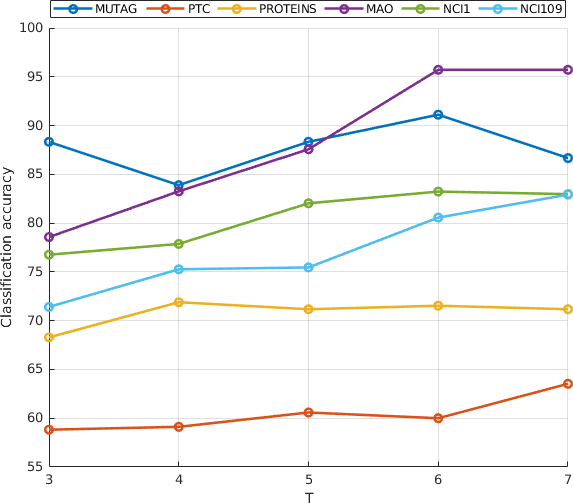}
\caption{Plot showing the classification accuracy obtained by SGE by varying the maximum number of edges from $3$ to $7$ on different datasets: MUTAG, PTC, PROTEINS, MAO, NCI1, NCI109.}
\label{fig:T_plot}
\end{figure}

Our algorithm is mainly controlled by three different parameters: (1) the \emph{number of levels} $L$ of the graph pyramid, (2) the \emph{reduction ratio} $R$ and (3) the maximum \emph{number of edges} $T$ of a graphlet. For illustrating how these three parameters control the performance of the system, first we plot the classification accuracy by varying the levels of the graph pyramid (see~\fig{fig:plots_levels}), reduction ratio (see~\fig{fig:plots_reduction}) and $T$ (see~\fig{fig:T_plot}). Here it is worth mentioning that for the sake of simplicity, for each level we just consider the maximum accuracy obtained by any configuration mentioned in~\sect{ssec:hembed}. From~\fig{fig:plots_levels}, we can observe that for all the datasets, considering a second level together with the base graph increases the classification accuracy. However, the successive inclusion of hierarchical levels does not always increase the performance. It has been observed that for smaller graphs (with less number nodes and edges; \eg, the graphs from MUTAG), the further inclusion of hierarchical abstraction decreases the performance of the system; this means that for smaller graphs a higher level abstraction can introduce noise or distortion. The reduction ratio $R$ directly decides the number of clusters in a given level, and hence the number of nodes in the next higher level of the hierarchy. For example, $R=1$ indicates that the number of clusters should remain the same with the number of nodes, while $R=2$ indicates that the number of clusters should be half the number of nodes in that level. \fig{fig:plots_reduction} shows the behaviour of our method with different values of $R$ while we have fixed $L=2$. From these plots, one must observe that $R$ is completely dependant on the datasets irrespective of the size of graphs they contain. For PTC, PROTEINS, and MAO datasets, the performance mostly increases with the increase of $R$, while for MUTAG, it improves until $R=2$, and then it decreases for all hierarchical configurations. For MAO dataset, all the hierarchical configurations behave exactly in the same way with the increase of $R$, which might be because the smaller sized graphs on which the contribution of different hierarchical configuration is indistinguishable.

In~\fig{fig:T_plot}, we show the performance trend on six datasets (i.e. MUTAG, PTC, PROTEINS, NCI1, and NCI109) only with the SGE algorithm, which is the baseline graph embedding technique that we considered. The hierarchical configurations are not considered in this case because they have different graphlet size in different hierarchical levels, so understanding their behaviour would have been complicated. From~\fig{fig:T_plot}, it is clear that increasing $T$ mostly improves the performance of the system on all the datasets. Albeit, there are some exceptions (\eg, for PTC dataset, $T=6$), which suggests that graphlets with $T$ edges are less informative for that particular graph dataset.

\subsection{Discussion on the stochasticity of the algorithm}

It is important to note that our proposed algorithm is stochastic in nature because of the involvement of the stochastic graphlet sampling and the subsequent graph embedding procedure. The graphlet sampling engaged here uniformly samples graphlets from a given population of graphs, and by the law of large numbers, this sampling guarantees that the empirical distribution of graphlets is asymptotically close to the actual distribution~\cite{Przulj2007}. For demonstrating the fact that the stochastic behaviour of our algorithm does not heavily impact on the experimental results, we repeated the last experiment on all the datasets considered for $10$ iterations, and in each iteration, we randomly seeded the sampling algorithm. The mean and standard deviation of the classification accuracy obtained for each dataset is reported in~\tab{t:pr-iter}. The mean accuracies reported in the table are quite close to the ones reported in~\tab{t:pr}, and the standard deviations are comparatively low (all of them are less than $1.0$). This suggests that the proposed graph embedding technique, although employed a stochastic process, is consistent in terms of performance.

\begin{table}[htb]
\centering
\caption{Mean and standard deviation of the accuracies obtained by repeating the classification task on the AIDS, GREC, COIL-DEL and HistoGraph datasets for $10$ iterations. Here the mean accuracies consistent with the ones in~\tab{t:pr} and the low standard deviations show that the proposed graph embedding is not sensitive to the stochasticity involved in the algorithm. The best results obtained on a dataset is specified by bold face.}
\label{t:pr-iter}
\resizebox{\columnwidth}{!}{
\begin{tabular}{l c c c c c c c c c}
\toprule
\multirow{2}{*}{\textbf{Methods}} & \multirow{2}{*}{\textbf{AIDS}} & \multirow{2}{*}{\textbf{GREC}} & \multirow{2}{*}{\textbf{COIL-DEL}} & \multicolumn{6}{c}{\textbf{HistoGraph}} \\
&  &  &  & \textbf{Keypoint} & \textbf{Grid-NNA} & \textbf{Grid-MST} & \textbf{Grid-DEL} & \textbf{Projection} & \textbf{Split} \\
\midrule
\multirow{2}{*}{\textbf{HSGE (pyr.)}} & $\mathbf{98.74}$ & $99.36$ & $98.74$ & $78.98$ & $\mathbf{72.71}$ & $77.57$ & $74.79$ & $79.72$ & $81.04$ \\
 & $(\pm 0.13)$ & $(\pm 0.19)$ & $(\pm 0.21)$ & $(\pm 0.32)$ & $(\pm 0.10)$ & $(\pm 0.43)$ & $(\pm 0.62)$ & $(\pm 0.99)$ & $(\pm 0.84)$ \\
\multirow{2}{*}{\textbf{HSGE (gen. pyr.)}} & $98.12$ & $99.58$ & $98.49$ & $\mathbf{79.31}$ & $71.28$ & $78.05$ & $74.96$ & $79.94$ & $80.24$ \\
 & $(\pm 0.27)$ & $(\pm 0.23)$ & $(\pm 0.49)$ & $(\pm 0.52)$ & $(\pm 0.58)$ & $(\pm 0.47)$ & $(\pm 0.71)$ & $(\pm 0.18)$ & $(\pm 0.74)$ \\
\multirow{2}{*}{\textbf{HSGE (hier.)}} & $98.24$ & $99.04$ & $\mathbf{98.98}$ & $79.03$ & $70.51$ & $76.20$ & $\mathbf{75.47}$ & $80.39$ & $80.38$ \\
 & $(\pm 0.36)$ & $(\pm 0.16)$ & $(\pm 0.60)$ & $(\pm 0.20)$ & $(\pm 0.55)$ & $(\pm 0.40)$ & $(\pm 0.86)$ & $(\pm 0.17)$ & $(\pm 0.21)$ \\
\multirow{2}{*}{\textbf{HSGE (exhaust.)}} & $\mathbf{98.74}$ & $\mathbf{99.64}$ & $98.84$ & $79.01$ & $71.96$ & $\mathbf{78.28}$ & $74.79$ & $\mathbf{80.82}$ & $\mathbf{81.53}$ \\
 & $(\pm 0.21)$ & $(\pm 0.80)$ & $(\pm 0.17)$ & $(\pm 0.70)$ & $(\pm 0.10)$ & $(\pm 0.97)$ & $(\pm 0.01)$ & $(\pm 0.46)$ & $(\pm 0.94)$ \\
\bottomrule
\end{tabular}}
\end{table}

\section{Conclusions}
\label{sec:concl}

In this paper we have proposed to enhance the information encoded in graph embeddings by means of hierarchical representations. We have experimentally validated that the abstract information is able to improve the graph classification performance.
The embedding function is based on a stochastic sampling of graphlets to obtain the graphlet distribution within the graph. Graphlets of different sizes are considered to allow a change on the node context. Moreover, the hashing functions are used to identify graphlets in an efficient way. Event though considering different size graphlets provides robustness in terms of graph distortions, they still provide local information when we consider larger graphs. Therefore, building a graph hierarchy allows to increase the graphlet context without increasing the time needed for identifying the graphlet. In this work, we have carefully validated the performance of our approach in different application scenarios, showing that we outperform the state-of-the-art approaches in the graph classification task using an SVM as a classifier.

Further research will focus on improving the hierarchical graph construction. Even though the Girvan-Newman algorithm is able to exploit the desired properties of the graph, creating clusterings that allow to create good abstractions, their time complexity is a drawback that should be studied when considering large graphs.

\section*{Acknowledgments}

This work has been partially supported by the European Union's research and innovation program under the Marie Sk\l{}odowska-Curie grant agreement No. 665919 (H2020-MSCA-COFUND-2014:665919), the Spanish projects RTI2018-102285-A-I00 and RTI2018-095645-B-C21, the FPU fellowship FPU15 / 06264 from the Spanish Ministerio de Educaci\'on, Cultura y Deporte, the Ramon y Cajal Fellowship RYC-2014-16831, and the CERCA Program / Generalitat de Catalunya. Anjan Dutta was a Marie-Curie Fellow (under the P-SPHERE project) at the Computer Vision Center of Barcelona, where most of the work was done and the paper was written.

\bibliographystyle{spmpsci}      
\bibliography{bibliography}   

\end{document}